\documentclass{article}

\usepackage{subcaption}
\usepackage{arxiv}
\usepackage{amsmath,amssymb}
\usepackage{graphicx}
\usepackage{hyperref}       
\def \method{PAGER}

\title{PAGER: Progressive Attribute-Guided Extendable Robust Image Generation}

\author{
  Zohreh Azizi \\
  Media Communications Lab\\
  University of Southern California\\
  Los Angeles, CA, USA \\
  \texttt{zazizi@usc.edu} \\
   \And
  C.-C. Jay Kuo \\
  Media Communications Lab\\
  University of Southern California\\
  Los Angeles, CA, USA \\
  \texttt{cckuo@sipi.usc.edu} \\
}

\begin{document}
\maketitle

\begin{abstract}
This work presents a generative modeling approach based on successive
subspace learning (SSL). Unlike most generative models in the
literature, our method does not utilize neural networks to analyze the
underlying source distribution and synthesize images. The resulting
method, called the progressive attribute-guided extendable robust image
generative (\method{}) model, has advantages in mathematical
transparency, progressive content generation, lower training time,
robust performance with fewer training samples, and extendibility to
conditional image generation. \method{} consists of three modules: core
generator, resolution enhancer, and quality booster. The core generator
learns the distribution of low-resolution images and performs
unconditional image generation. The resolution enhancer increases image
resolution via conditional generation.  Finally, the quality booster
adds finer details to generated images.  Extensive experiments on MNIST,
Fashion-MNIST, and CelebA datasets are conducted to demonstrate
generative performance of \method{}. 
\end{abstract}


\section{Introduction}\label{sec:intro}

Unconditional image generation has been a hot research topic in the last
decade. In image generation, a generative model is trained to learn the
image data distribution from a finite set of training images. Once
trained, the generative model can synthesize images by sampling from the
underlying distribution. 

GANs have been widely used for unconditional image generation with
impressive visual quality in recent years
\cite{goodfellow2014generative}. Despite the evident advantages of GANs,
their training is a non-trivial task: GANs are sensitive to training
hyperparameters and generally suffer from convergence issues
\cite{hoshen2019non}. Moreover, training GANs requires
large-scale GPU clusters and an extensive number of training data.
\cite{liu2020towards}. Limited training data usually cause the
discriminator to overfit and the training to diverge
\cite{karras2020training}. These concerns have led to the development of
improved GAN training methods \cite{gulrajani2017improved}, techniques
for stabilized training with fewer data \cite{liu2020towards,
karras2020training}, or non-adversarial approaches
\cite{hoshen2019non}. Yet, the great majority of existing generation
techniques utilize deep learning (DL), a method for learning deep neural
networks, as the modeling backbone. 

A neural network is typically trained using a large corpus of data over
long episodes of iterative updates. Therefore, training a neural network
is often a time-consuming and data-hungry process. To ensure the
convergence of deep neural networks (DNNs), one has to carefully select
(or design) the neural network architecture, the optimization objective
(or the loss) function, and the training hyper-parameters.
Some DL-based generative models like GANs are often
specifically engineered to perform a certain task. They cannot be easily
generalized to different related generative applications. For example,
the architectures of these neural networks for unconditional image
generation have to be re-designed for image super-resolution or
attribute-guided image generation. Last but not the least, due to the
non-linearity of neural networks, understanding and explaining their
performance is a standing challenge. 

To address the above-mentioned concerns, this paper presents an
alternative approach for unconditional image generation based on
successive subspace learning (SSL) \cite{kuo2016understanding,
kuo2017cnn, kuo2018data, kuo2019interpretable}. The resulting method,
called progressive attribute-guided extendable robust image generative
(\method{}) model, has several advantages, including mathematical
transparency, progressive content generation, lower training time,
robust performance with fewer training samples, and extendibility to
conditional image generation. 

\method{} consists of three modules: 1) core generator, 2) resolution
enhancer, and 3) quality booster. The core generator learns the
distribution of low-resolution images and performs unconditional image
generation.  The resolution enhancer increases image resolution via
conditional generation. Finally, the quality booster adds finer details
to generated images. 

To demonstrate the generative performance of \method{}, we conduct
extensive experiments on MNIST, Fashion-MNIST, and CelebA datasets.  We
show that \method{} can be trained in a fraction of the time required
for training DL based models and still achieve a similar generation
quality. We then demonstrate the robustness of \method{} to the training
size by reducing the number of training samples. Next, we show that
\method{} can be used in image super resolution, high-resolution image
generation, and attribute-guided face image generation. In particular,
the modular design of \method{} allows us to use the conditional
generation modules for image super resolution and high-resolution image
generation. The robustness of \method{} to the number of training
samples enables us to train multiple sub-models with smaller subsets of
data. As a result, \method{} can be easily used for attribute-guided
image generation. 

The rest of this paper is organized as follows. Related work is reviewed
in Sec.~\ref{sec:related}. The \method{} method is proposed in
Sec.~\ref{sec:method}. Experimental results are reported in
Sec.~\ref{sec:experiments}. Extendability and applications of \method{}
are presented in Sec.~\ref{sec:applications}. Finally, concluding
remarks and possible future extensions are given in
Sec.~\ref{sec:conclusion}. 

\section{Related Work}\label{sec:related}

\subsection{DL-based Image Generative Models}\label{subsec:DL}

DL-based image generative models can be categorized into two main
classes: adversarial-based and non-adversarial-based models.  GANs
\cite{goodfellow2014generative} are adversarial-based generative models
that consist of a generator and a discriminator. The training procedure
of a GAN is a min-max optimization where the generator learns to
generate realistic samples that are not distinguishable from those in
the original dataset and the discriminator learns to distinguish between
real and fake samples. Once the GAN model is trained, the generator
model can be used to draw samples from the learned distribution.
StyleGANs have been introduced in recent years. They exploit
the style information, leading to better disentangability and
interpolation properties in the latent space and enabling better control
of the synthesis \cite{karras2019style, karras2020analyzing,
karras2021alias}.

Examples of non-adversarial DL-based generative models include
variational auto-encoders (VAEs) \cite{kingma2013auto}, flow-based
models \cite{dinh2014nice, dinh2016density}, GLANN \cite{hoshen2019non},
and diffusion-based models \cite{dhariwal2021diffusion, ho2022cascaded}.
VAEs have an encoder/decoder structure that learns variational
approximation to the density function. Then, they generate images from
samples of the Gaussian distribution learnt through the variational
approximation. An improved group of VAEs called
Vector-Quantized VAEs (VQ-VAE) can generate outputs of higher quality.
In VQ-VAEs, the encoder network outputs discrete codes and the prior is
learnt instead of being static \cite{razavi2019generating,
van2017neural}. Flow-based methods apply a series of invertible
transformations on data to transform the Gaussian distribution into a
complex distribution. Following the invertible transformations, one can
generate images from the Gaussian distribution. GLANN
\cite{hoshen2019non} employs GLO \cite{bojanowski2017optimizing} and
IMLE \cite{li2018implicit} to map images to the feature and the noise
spaces, respectively. The noise space is then used for sampling and
image generation.  Recently, diffusion-based models are developed for
image generation. During the training process, they add noise to images
in multiple iterations to ensure that the data follows the Gaussian
distribution ultimately. For image generation, they draw samples from
the Gaussian distribution and denoise the data in multiple gradual steps
until clean images show up. 

Despite impressive results of DL-based generative models, they are
mathematically not transparent due to their highly non-linear
functionality. Furthermore, they are often susceptible to unexpected
convergence problems \cite{hoshen2019non}, long training time, and
dependency on large training dataset size. As we show in our
experiments, \method{} addresses the aforementioned concerns while
maintaining the quality of the images generated by DL-based techniques. 

\subsection{Unconditional and Conditional Image Generation}

In unconditional image generation, sample images are drawn from an
underlying distribution without any prior assumption on the images to be
generated. In conditional image generation, samples are generated under
a certain assumption. One example of the latter is the generation of a
high-resolution image given a low-resolution image. The proposed
\method{} method contains both unconditional and conditional image
generation techniques. Its core generator module employs the
unconditional image generation technique. Its resolution enhancer and
quality booster modules perform conditional image generation. Although
\method{} is an unconditional image generator by itself, it can be
easily extended to conditional image generation with rich applications.
We will elaborate this point with three examples, namely,
attribute-guided face image generation, image super resolution, and
high-resolution image generation. Each task is elaborated below.

\textbf{Attribute-guided face image generation:} For a set of required
facial attributes, the goal is to generate face images that meet the
requirements. \cite{lu2018attribute} performs attribute-guided face
image generation using a low-resolution input image. It modifies the
original CycleGAN \cite{zhu2017unpaired} architecture and its loss
functions to take conditional constraints during training and inference.
In \cite{kowalski2020config}, synthetic labeled data are used to
factorize the latent space into sections which associate with separate
aspects of face images. It designs a VAE with an additional attribute
vector to specify the target part in the factorized latent space.
\cite{qian2019make} proposes to learn a geometry-guided disentangled
latent space using facial landmarks to preserve generation fidelity. It
utilizes a conditional VAE to sample from a combination of
distributions. Each of them corresponds to a certain attribute. 

\textbf{Image super-resolution:} The problem aims at generating a
high-resolution image that is consistent with a low-resolution image
input. One solution is the example-based method
\cite{freeman2002example}. Others include auto-regressive models and
normalized flows \cite{van2016conditional,parmar2018image,
yu2020wavelet}. Quite a few recent papers adopt the DL methodology
\cite{dong2014learning}. Another line of work treats super-resolution as
a conditional generation problem, and utilize GANs or diffusion-based
models as conditional generative tools which use low-resolution images
as the generation condition \cite{ledig2017photo,chen2018fsrnet,
saharia2021image}. 

\textbf{Progressive generation of very-high-resolution Images:}
Generation of a very-high-resolution image of high quality is
challenging and treated as a separate research track.  A
common solution is to take a progressive approach in training and
generation to maintain the model stability and generation quality.
There exist both GAN-based and diffusion-based very-high-resolution
image generation solutions \cite{karras2017progressive, ho2022cascaded}. 

Our \method{} method can be trained for unconditional image generation
as well as for conditional image generation such as attribute-guided
face image generation and image super-resolution. In principle, it can
also be used for progressive generation of very-high-resolution images.
Our \method{} serves as a general framework that can bridge different
generation models and applications. 

\subsection{Successive Subspace Learning (SSL)}\label{subsec:SSL}

In order to extract abstract information from visual data, spectral or
spatial transforms can be applied to images. For example, the Fourier
transform is used to capture the global spectral information of an image
while the wavelet transform can be exploited to extract the joint
spatial/spectral information. Two new transforms, namely, the Saak
transform \cite{kuo2018data} and the Saab transform
\cite{kuo2019interpretable}, were recently introduced by Kuo {\em et
al.} \cite{kuo2016understanding, kuo2017cnn, kuo2018data,
kuo2019interpretable} to capture joint spatial/spectral features. These
transforms are derived based on the statistics of the input without
supervision. Furthermore, they can be cascaded to find a sequence of
joint spatial-spectral representations in multiple scales, leading to
Successive Subspace Learning (SSL). The first implementation of SSL is
the PixelHop system \cite{chen2020pixelhop}, where multiple stages of
Saab transforms are cascaded to extract features from images. Its second
implementation is PixelHop++, where channel-wise Saab transforms are
utilized to achieve a reduced model size while maintaining an effective
representation \cite{chen2020pixelhop++}. An interesting
characteristic of the Saab transform that makes SSL a good candidate for
generative applications is that it is invertible. In other words, the
SSL features obtained by multi-stage Saab transforms can be used
to reconstruct the original image via the inverse SSL, which is formed
by multi-stage inverse Saab transforms. Once we learn the Saab
transform from training data, applying the inverse Saab transform in inference would be trivial.\footnote{\url{https://github.com/zohrehazizi/torch_SSL}}

SSL has been successfully applied to many image processing and computer
vision applications \cite{rouhsedaghat2021successive}. Several examples include unconditional image generation \cite{lei2022genhop, lei2020nites, lei2021tghop}, point cloud analysis \cite{zhang2020pointhop, zhang2020pointhop++, zhang2020unsupervised, kadam2020unsupervised, zhang2021gsip, kadam2021gpco, kadam2021r}, fake image detection
\cite{chen2021defakehop, chen2021geo, chen2022defakehop++,
zhu2021pixelhop}, face recognition \cite{rouhsedaghat2021facehop,
rouhsedaghat2021low}, medical diagnosis \cite{liu2021voxelhop,
monajatipoor2021berthop}, low light enhancement \cite{azizi2020noise},
anomaly detection \cite{zhang2021anomalyhop}, to name a few.  Inspired
by the success of SSL, we adopt this methodology in the design of a new
image generative model as elaborated in the next section. 

\subsection{SSL-based Image Generative Models}\label{subsec:GenHop}

GenHop \cite{lei2022genhop} is the contemporary SSL-based image
generative model in literature. GenHop utilizes SSL for feature
extraction. It applies independent component analysis (ICA) and
clustering to obtain clusters of independent feature components at the
last stage of SSL.  Then, it finds a mapping between the distribution of
ICA features and Guassian distributions.  In this work, we do not
perform ICA but model the distribution of SSL features via GMMs
directly. As compared to GenHop, our approach offers several attractive
features. First, it has lower computational complexity and demands less
memory. Second, our method offers a progressive and modular image
generation solution. It is capable of conditional and attribute-guided
image generation. It can also be easily extended to other generative
applications such as super-resolution or high-resolution image
generation.

\section{Proposed \method{} Method}\label{sec:method}

The \method{} method is presented in this section. First, our research
motivation is given in Sec. \ref{subsec:motivation}. Then, an overview
on \method{} and its three modules are described in Sec.
\ref{subsec:overview}.  Finally, our attribute-guided face image
generation is elaborated in Sec. \ref{subsec:attributes}. 

\subsection{Motivation}\label{subsec:motivation}

A generative model learns the distribution of the training data in the
training phase. During the generation phase, samples are drawn from the
distribution as new data.  To improve the accuracy of generative image
modeling, gray-scale or color images should be first converted into
dimension-reduced latent representations. After converting all training
images into their (low-dimensional) latent representation, the
distribution of the latent space can be approximated by a multivariate
Gaussian distribution.  For learning the latent representation, most
prior work adopts GAN-, VAE-, and diffusion-based generative models;
they train neural networks that can extract latent representations from
an image source through a series of nonlinear transformations.
Similarly, we need to learn such a transformation from the image space
to the latent representation space. 

\begin{figure*}[t]
\includegraphics[width=\linewidth]{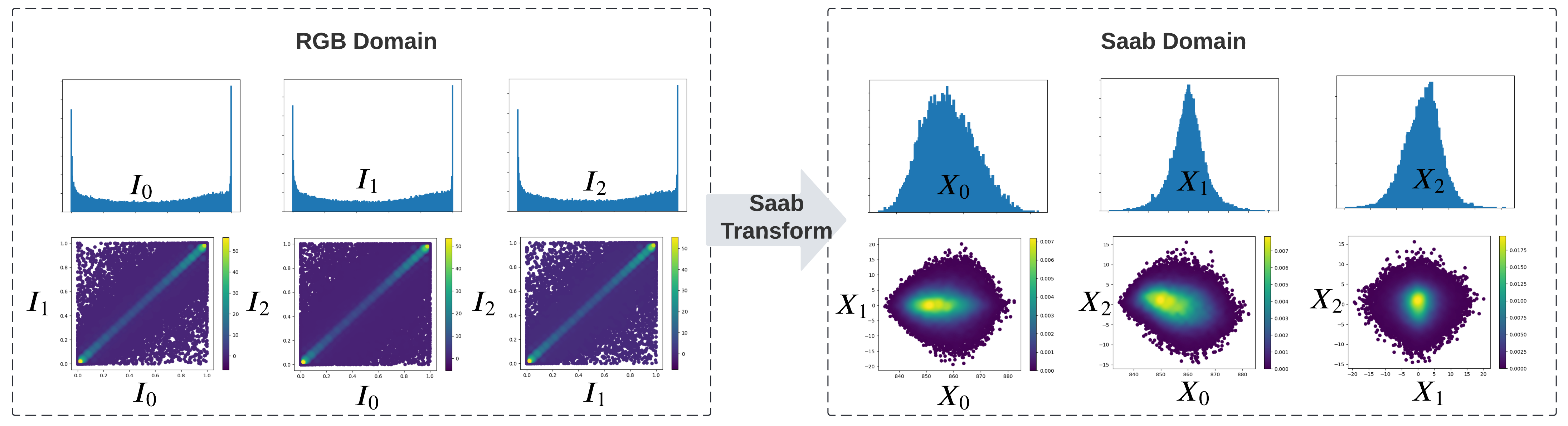}
\caption{Example distributions from RGB pixels (left block) and Saab
transforms (right block). The top figures correspond to single vector
dimensions ($I_0\dots I_2$ in RGB and $X_0\dots X_2$ in Saab domains). The
bottom figures correspond to joint distributions. Distributions are
extracted from the first three components of CelebA
images.}\label{fig:gmm}
\end{figure*}

In this work, we utilize an SSL pipleline, rather than neural networks,
to achieve the transformation to the latent representation space. The
SSL pipeline consists of consecutive Saab transforms. In
essence, it receives an image, denoted by $I\in \mathbb{R}^{w\times
h\times c}$, and converts it into a latent feature vector, denoted by
$X\in \mathbb{R}^n$, where $w$, $h$ and $c$ are the pixel numbers of the
width, height and color channels of an image while $n$ is the dimension
of the latent vector. For the remainder of this paper, we refer to the
latent space obtained by SSL as the \textit{core space}. The Saab
transform utilizes mean calculation and PCA computation to extract
features from its input.  Due to the properties of PCA, the $i$-th and
$j$-th components in the core space are uncorrelated for $i\neq j$. This
property facilitates the use of Gaussian priors for generative model
learning over the core space. 

Fig.~\ref{fig:gmm} illustrates the distributions of input image pixels
($I$) and Saab outputs ($X$). In this example, we plot the distributions
of the first, second and third components of $I$ (i.e., the RGB values
of the upper-left pixel of all source images) and $X$ (i.e., the Saab
transform coefficients).  The RGB components are almost uniformly
distributed in the marginal probability. They are highly correlated as
shown in the plot of joint distributions.  In contrast, Saab
coefficients are close to the Gaussian distribution and they are nearly
uncorrelated. While the distributions of one- and two-dimensional
components of $X$ are very close to Gaussians, the distribution of
higher-dimensional vectors might not be well modeled by one multivariate
Gaussian distribution. For this reason, we employ a mixture of Gaussians
to represent the distribution of the core space. 

\begin{figure*}[t]
\includegraphics[width=\linewidth]{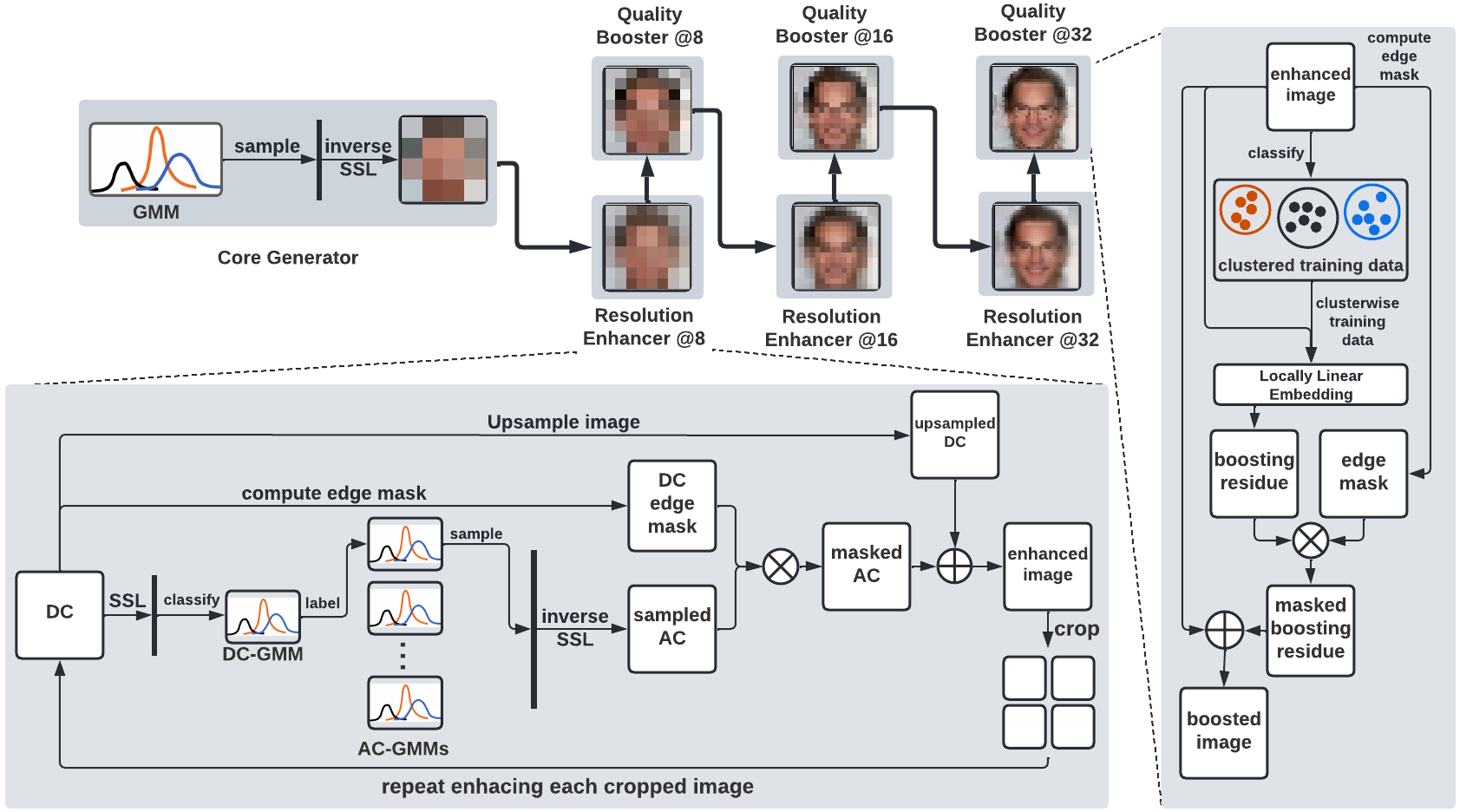}
\caption{Overview of \method{} generation method.}\label{fig:overview}
\end{figure*}

\subsection{System Overview}\label{subsec:overview}

An Overview of the \method{} generation method is shown in
Fig.~\ref{fig:overview}. \method{} is an unconditional generative model with a progressive approach in image generation. It starts with unconditional generation in a low-resolution regime, which is performed by the core generator. Then, it sequentially increases the image resolution and quality through a cascade of two conditional generation modules: the resolution enhancer and the quality booster.

\subsubsection{Module 1: Core Generator}\label{sec:generation}

The core generator is the unconditional generative module in \method{}. Its goal is to generate low-resolution (e.g., $4\times4\times 3$) color images. This module is trained with images of shape $2^d\times 2^d\times 3$ (e.g., $d=2$). It applies consecutive Saab transforms on input images $\{I_i\}_{i=1}^{M}$ using PixelHop++ structure \cite{chen2020pixelhop++}, ultimately converting images into $n$-dimensional vectors $X\in \mathbb{R}^n$ ($n=2^d\times2^d\times3$) in core space. The goal of the core generator is to learn the distribution of $\{X_i\}_{i=1}^{M}$. We use $\mathcal{X}$ to denote a random variable within $\{X_i\}_{i=1}^{M}$, representing observed samples in core space. Let $P(\mathcal{X})$ be the underlying distribution of $\mathcal{X}\in \mathbb{R}^n$. The generation core $G$ attempts to approximate the distribution $P(\mathcal{X})$ with a distribution $G(\mathcal{X})$. 

DL-based methods utilize iterative end-to-end optimization of neural networks to achieve this objective. In \method{}, we model the underlying
distribution of the core space using the Gaussian Mixture Model (GMM), which is highly efficient in terms of training time. This is feasible since we use SSL to decouple random variables, which we illustrated in Sec. \ref{subsec:motivation}. The conjunction of multi-stage Saab (SSL) features and GMMs can yield a highly accurate density modeling. Formally, the GMM approximation of $G(\mathcal{X})$ is defined as follows:

\begin{equation}\label{eq:GMM}
G(\mathcal{X})=\sum_{k=1}^{K} p_k \mathcal{N}(\mathcal{X},\mu_k,\Sigma_k),
\end{equation}
where $\mathcal{N}(\mathcal{X},\mu_k,\Sigma_k)$ is a multi-variate
normal distribution with mean $\mu_k$ and diagonal covariance matrix
$\Sigma_k$, and $p_k$ is a binary random variable. We have $p_k=1$ with
probability $P_k$, $p_k=0$ with probability $(1-P_k)$ and
$\sum_{k=1}^{K} P_k=1$. In other words, only one of the $K$ Gaussian
models will be selected at a time, and the probability of selecting 
the $k$-th Gaussian model is $P_k$ in such a GMM. The parameters of the 
GMM can be determined using the Expectation Maximization (EM) algorithm
~\cite{reynolds2009gaussian}.  Once such a GMM model is obtained, one
can draw a sample, $X$, randomly and proceed to Modules 2 and 3. 

The need for Modules 2 and 3 is explained below. $G(\mathcal{X})$ is
learned from observations $X_i$, $i=1 \cdots M$. When the dimension,
$n$, of the core space is large, estimating $G(\mathcal{X})$ becomes
intractable and the approximation accuracy of GMM would drop. For this
reason, the unconditional generation process is constrained to a
low-dimensional space. Then, we employ conditional generative models (modules 2 and 3) to
further increase image resolution and quality.

\subsubsection{Module 2: Resolution Enhancer}\label{subsec:module_2}

We represent image $I_d$ as the summation of its DC and AC components:
\begin{eqnarray}\label{eq:DC_AC_decomposition}
I_d &=& DC_d + AC_d, \\
DC_d &=& U(I_{d-1}),
\end{eqnarray}
where ${I_{d}}$ is an image of size ${2^{d}\times 2^{d}}$, $U$ is the
Lanczos image interpolation operator, $DC_d$ is the interpolated image
of size ${2^{d}\times 2^{d}}$ and $AC_d$ is the residual image of size
$2^d \times 2^d$.  The above decoupling of DC and AC components of an
image allows to define the objective of the resolution enhancer. It
aims to generate the residual image $AC_d$ conditioned on $DC_d$. In
Fig.~\ref{fig:overview}, a multi-stage cascade of resolution enhancers
is shown. The detail of a representative resolution enhancer is
highlighted in the lower subfigure. 

To train the resolution enhancer, we first decouple the DC and AC of
training samples. Then, we extract SSL features from the DC and build a
GMM model with $K$ components, denoted by $G_{DC}$. By this method, we
learn a distribution of the DC at a certain image resolution. Note that
each DC from a training image belongs to one of the Gaussian models in
$G_{DC}$. Therefore, DCs (and their associated AC) are clustered into
$K$ classes using $G_{DC}$.  We gather the AC of each class and build a
corresponding GMM, denoted by $G_{AC,k}$ where $k\in\{1, \cdots, K\}$.
In total, we learn $K+1$ GMMs: $\{G_{DC}, \, G_{AC,1} \, \dots \,
G_{AC,K}\}$. 

At the test time, the resolution enhancer receives the low resolution
image $I_{d-1}$, and upsamples it to obtain the interpolated DC, i.e.,
$DC_d = U(I_{d-1})$.  Then, the resolution enhancer converts the DC to
its SSL features and classifies it into one of the $K$ clusters using
$G_{DC}$. Mathematically, we have
\begin{eqnarray}
X_{DC} &=& \mbox{SSL} (DC_d), \\
y &=& \mbox{arg}_k \max {\{\mathcal{N}(X_{DC}, \mu_k, \Sigma_k)\}_{k=1}^{K}},
\end{eqnarray}
where $\mathcal{N}(X_{DC}, \mu_k, \Sigma_k)$ is the probability score of
$X_{DC}$ according to the $k$-th component of $G_{DC}$, and the
classification label $y$ is the maximizer index.  In other words, the
resolution enhancer identifies a cluster of samples that are most
similar to $DC_{d}$. Next, the resolution enhancer draws a sample from
the $AC$ distribution corresponding to class $y$:
\begin{equation}
X_{AC} \sim G_{AC,y}(\mathcal{X}_{AC}).
\end{equation}
With the above two-step generation, the resolution enhancer generates
$X_{AC}$ conditioned on $X_{DC}$. Afterwards, $X_{AC}$ is converted to
the RGB domain using the inverse SSL transform:
\begin{equation}\label{eq:ac_inverse}
AC_d = \mbox{SSL}^{-1}(X_{AC}).
\end{equation}

The computed AC component is masked and added to the DC to yield
the higher resolution image via
\begin{eqnarray}\label{eq:mask}
I_d &=& DC_d+\widehat{AC}_d, \\
\widehat{AC}_d &=& M(DC_d)\odot AC_d,
\end{eqnarray}
where $M(DC_d)$ is a mask and $\odot$ denotes element-wise
multiplication.  The mask is derived from the edge information obtained
by the Canny edge detector~\cite{canny1986computational}. The masking
operation serves two objectives. First, it prevents details from being
added to smooth regions of the DC component. Second, it suppresses
unwanted noise.  Once $I_d$ is generated, it is cropped into four
non-overlapping regions, and each region goes through another resolution
enhancement process. The process is recursively applied to each
sub-region to further enhance image quality. In our experiments, we
continue the recursion until a cropped window size of $2\times 2$ is reached. 

\subsubsection{Module 3: Quality Booster}\label{subsec:module_3}

The right subfigure of Fig.~\ref{fig:overview} presents the quality
booster module. It follows the resolution enhancer by adding detail and
texture to the output of the resolution enhancer. It exploits the
locally linear embedding (LLE)~\cite{roweis2000nonlinear} scheme and
adds extra residue values that are missed by the resolution enhancer.
LLE is a well known method in building correspondence between two
components in image super resolution~\cite{chang2004super,
johnson2019billion} or image restoration~\cite{huang2016visible}.  To
design the quality booster, we decompose the training dataset, enhance
the DC component, and compute the residuals as follows:
\begin{eqnarray}
I_d & = & DC_d+AC_d, \\
E_d & = & \mbox{Enhancer}(DC_d), \\
R_d & = & I_d- E_d,
\end{eqnarray}
where ${I}_d$ represents a ${2^d\times 2^d}$ training image, ${E_d}$ is
the result of applying the enhancer module to the DC component of the
image, and $R_d$ is the residual image.  During training, the quality
booster stores $E_d^i$ and $R_d^i$, $i=1, \cdots, M$ from $M$ training
samples.  In generation, the quality booster receives image $E_d$ and
uses the LLE algorithm to estimate the residual image for image $E_d$
based on $E_d^i$ and $R_d^i$ from the training dataset. It approximates
the residual image with a summation of several elements within $R_d^i$.
Readers are referred to \cite{roweis2000nonlinear} for details of LLE 
computation. Similar to the enhancer module, the computed $R_d^i$ is
masked and added to $E_d$ to boost its quality. 

Although the LLE in the quality booster module uses training
data residues during inference, it does not affect the generation
diversity for two reasons. First, the quality booster only adds some
residual textures to the image. In other words, it has a sharpening
effect on edges. Since its role is limited to adding residuals and
sharpening, it does not have a significant role in adding or preventing
diversity. Second, the weight prediction mechanism of LLE provides a
method to combine various patch instances and obtain diverse patterns.

\subsection{Attribute-Guided Face Image Generation}\label{subsec:attributes}

In attribute-guided face image generation, the goal is to synthesize
face images that have certain properties. Let $A\in\{-1, +1\}^T$ denote
a set of $T$ binary attributes. The goal is to synthesize an image that
satisfies a query ${\bf q}\in\{-1, 0, +1\}^T$, where -1, 0, +1 denote
negative, don't care, and positive attributes. For instance, if the
attribute set is \{\textit{male}, \textit{smiling}\}, the query ${\bf
q}=[-1, +1]$ requests an image of a female smiling person, and the query
${\bf q}=[0,-1]$ request an image (of any gender) that is not smiling.

Without loss of generality, we explain the attribute-guided generation
process with $T=7$. The attributes selected from attribute labels in
CelebA dataset include `gender', `smiling', `blond hair', `black hair',
`wearing lipstick', `bangs' and `young'. Given these seven binary
attributes, there are $2^7=128$ subsets of data that correspond to each
unique set of selected attributes. However, some of the attribute
combinations might not be abundant in the training data due to the
existing correlation between the attributes. For instance, `wearing
lipstick', `bangs', and `gender' are highly correlated. Thus, instead of
considering all 128 combinations, we partition the attributes of
training data into $K$ subsets using k-means clustering (we set $K=10$
in our experiments).  Based on the attribute clusters, we create $K$
data subsets and train a separate \method{} model for each subset. 

\begin{figure*}[t]
\includegraphics[width=\linewidth]{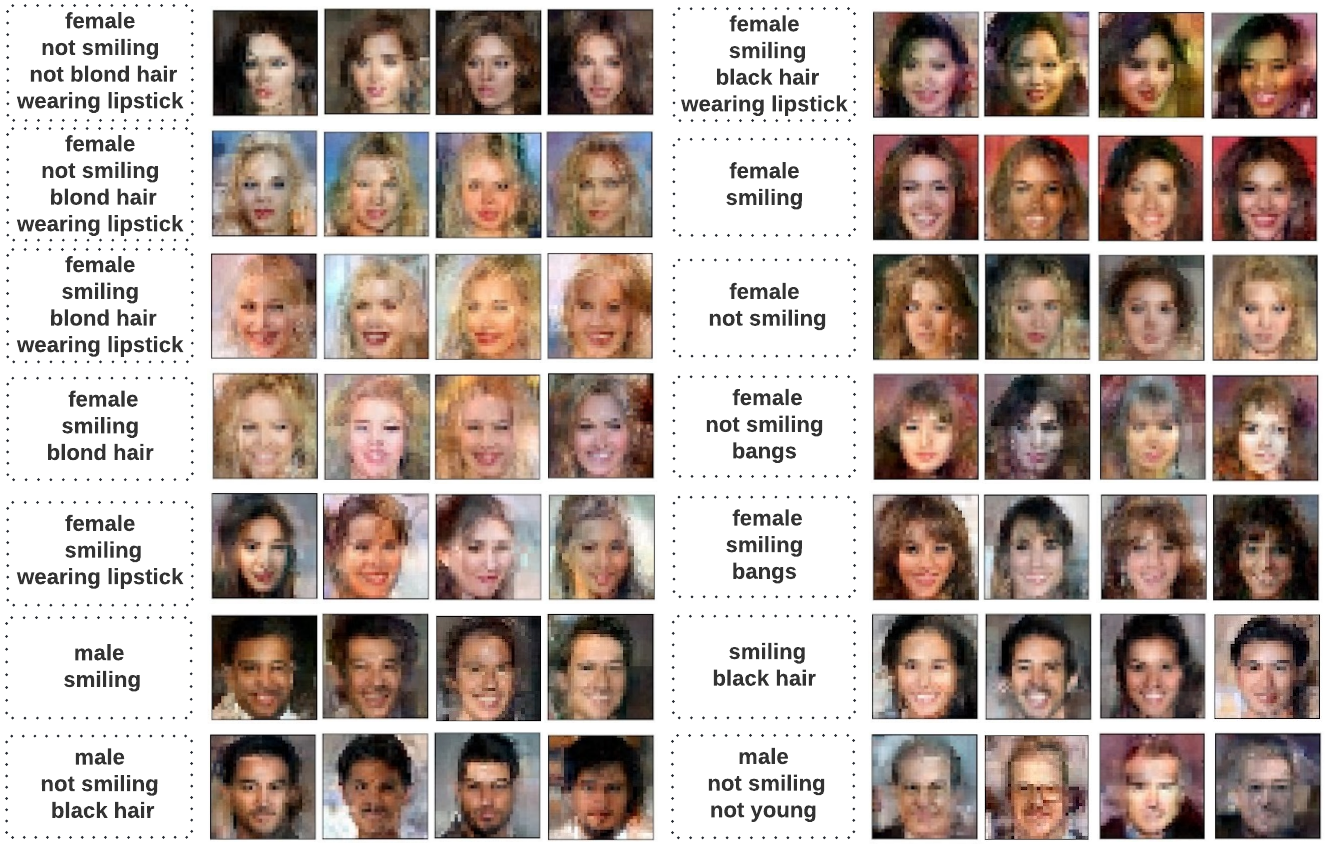}
\caption{Examples of attribute-guided generated images for CelebA with various attribute
combinations.}\label{fig:attributes}
\end{figure*}

At generation time, the goal is to synthesize a sample with a given
attribute set, ${\bf q}\in\{-1, 0, +1\}^7$.  To determine which of the 10
models best represents the requested attribute set, we compute the
Cosine distance of ${\bf q}$ to each of the cluster centers and select
the model that gives the minimum distance. Then, we draw samples from
the corresponding model.  Fig.~\ref{fig:attributes} shows generated
images corresponding to 15 different attribute vectors. We see that the
attribute-based generation technique can successfully synthesize images
with target attributes while preserving diversity and fidelity. 

\section{Experiments}\label{sec:experiments}

\subsection{Experimental Setup}

We perform experiments on three datasets: MNIST, Fashion-MNIST, and
CelebA. They are commonly used for learning unconditional image
generative models. We briefly explain the experimental settings of
\method{} for each dataset below. 

\textbf{CelebA.} The dataset is a set of colored human face images.
Suppose that there are $2^d\times 2^d$ pixels per image. To derive Saab
features and their distributions, we apply $d$-stage cascaded Saab
transforms. At each stage, the Saab filter has a spatial dimension of
$2\times2$ with stride $2$. The number of GMM components in the core
generator is $500$. The core generator synthesizes color images of size
$4\times 4$. Higher resolution images are generated conditioned on the
previous resolution with the resolution enhancer and the quality booster
modules in cascade ($4\times 4 \rightarrow 8\times 8 \rightarrow
16\times 16 \rightarrow 32\times 32$). The resolution enhancer has $100$
GMM components for the DC part and $3$ GMM components for the AC part at
each stage. LLE in the quality booster module is performed using $2$
nearest neighbors. 

\textbf{MNIST and Fashion-MNIST.} The two datasets contain gray-scale
images of digits and clothing items, respectively. The generation
pipeline for these datasets is similar to CelebA except that the core
generator synthesizes $16\times 16$ padded gray-scale images for each of the
$10$ classes. The $16\times 16$ images are converted to $32\times 32$
with a single stage of resolution enhancer and quality booster. Finally, they are cropped to $28\times 28$.

\begin{figure*}[t]
\centering
\includegraphics[width=\linewidth]{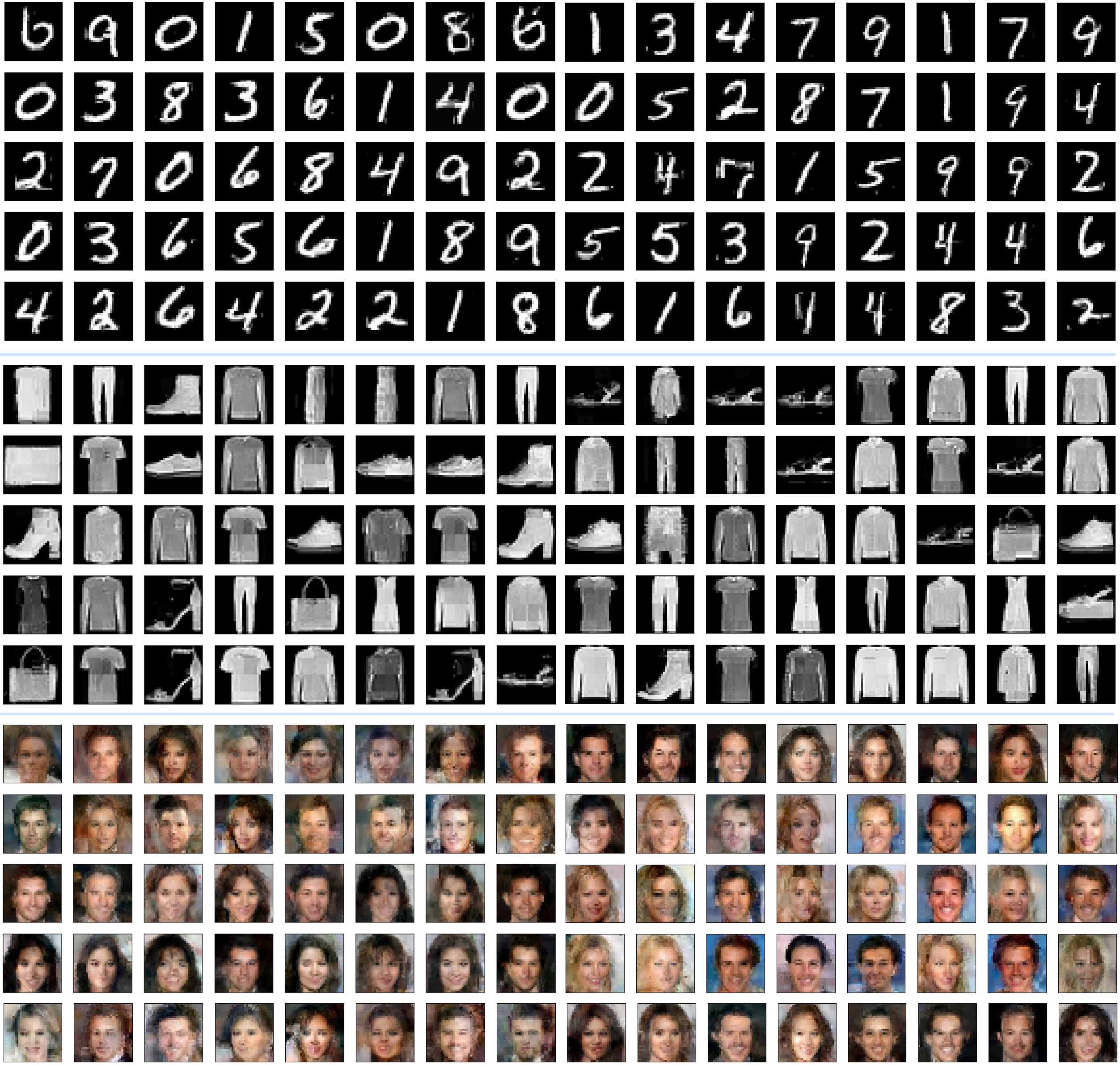}
\caption{Examples of \method{} generated images for MNIST (top),
Fashion-MNIST (middle), and CelebA (bottom) datasets.} \label{fig:mnist}
\end{figure*}

\subsection{Evaluation of Generated Image Quality}

{\bf Subjective Evaluation.} We show image samples of resolution $32
\times 32$ generated by \method{} for MNIST, Fashion-MNIST and CelebA in
Fig.~\ref{fig:mnist}. Generated images learned from MNIST represent the
structure of digits accurately and with rich diversity. Images generated
from Fashion-MNIST show diverse examples for all classes with fine
details and textures. Generated images for CelebA are semantically
meaningful and with fine and diverse details in skin tone, eyes, hair and
lip color, gender, hairstyle, smiling, lighting, and angle of view. 

\begin{figure}
    \centering
    \includegraphics[width=\columnwidth]{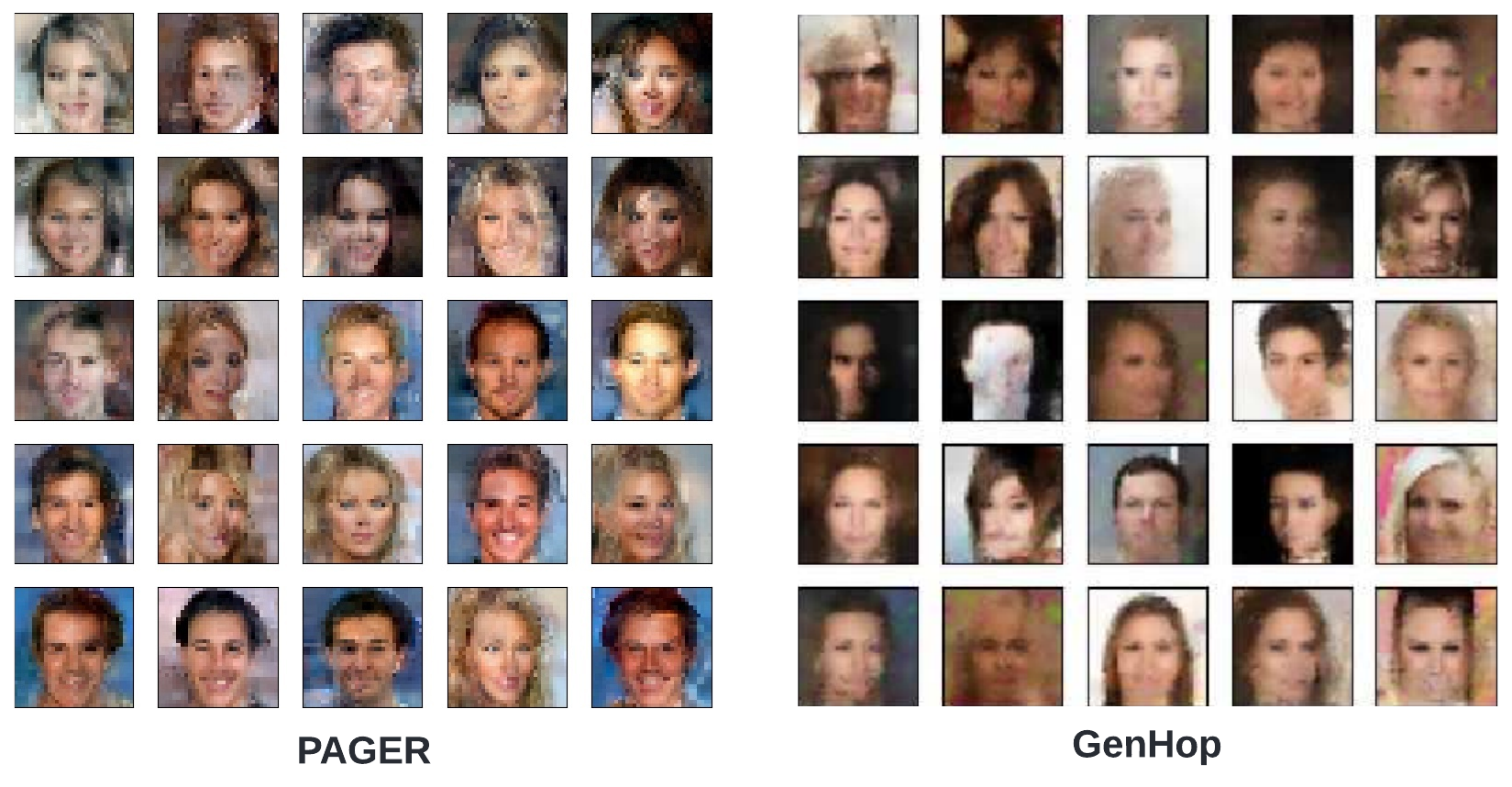}
    \caption{Example images generated by \method{} and GenHOP for the CelebA dataset.}
    \label{fig:CelebaComparisonGenhop}
\end{figure}

Fig.~\ref{fig:CelebaComparisonGenhop} compares generated
images by GenHop~\cite{lei2022genhop}, which is an earlier SSL-based
method, and our \method{{}} for the CelebA dataset. To be compatible
with GenHop, we perform comparison on generated images of resolution
$32\times 32$. As seen, images generated by \method{} are more realistic
with finer details than GenHop.

\begin{figure}
    \centering
    \includegraphics[width=\columnwidth]{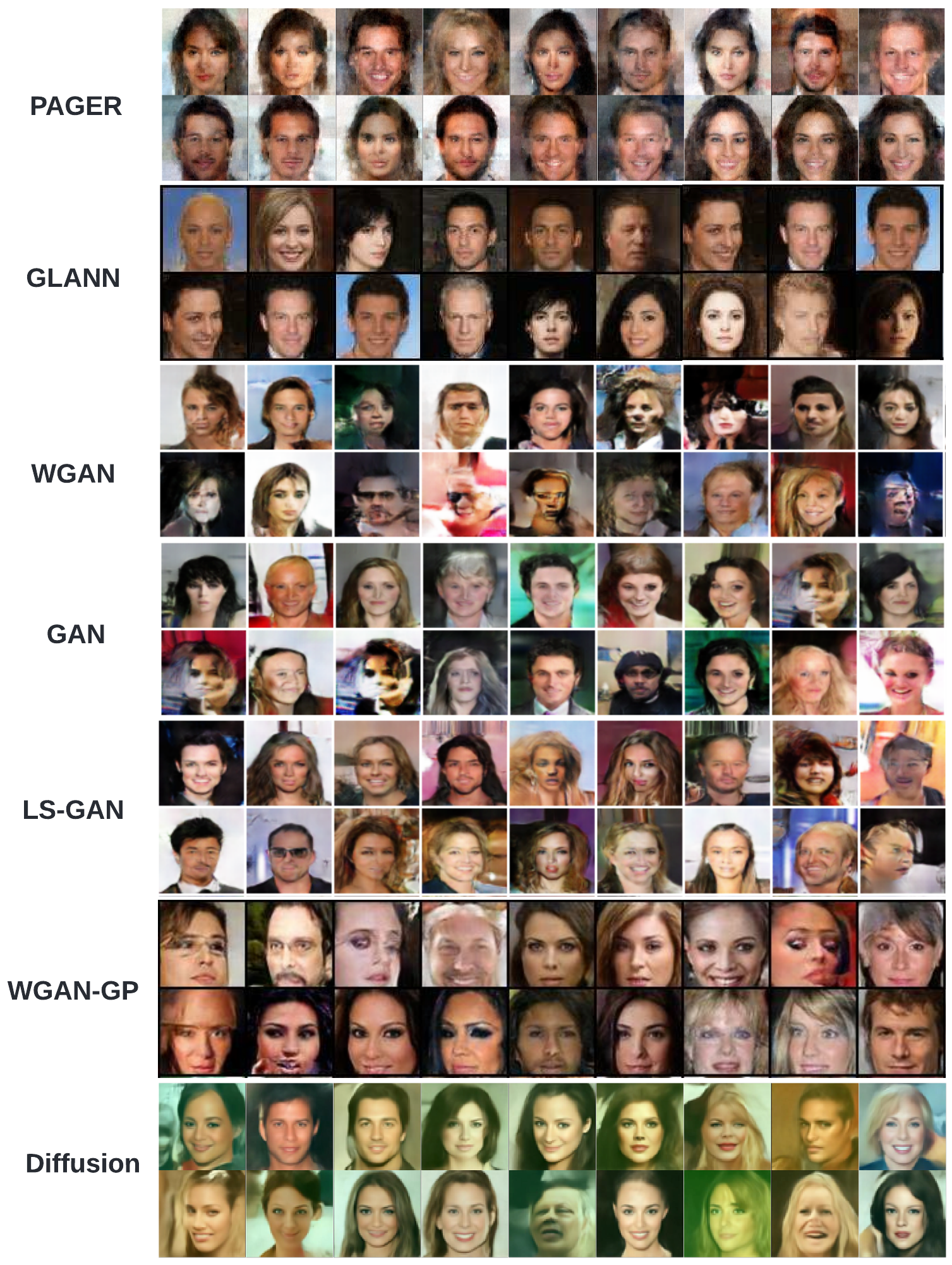}
\caption{Samples generated by \method{} and prior DL-based
generative models for the CelebA dataset.}\label{fig:CelebaComparisonDL}
\end{figure}

Next, we compare images generated by our method and those
obtained by prior DL-based generative models in Fig.
\ref{fig:CelebaComparisonDL}. We resort our comparison to
GAN~\cite{goodfellow2014generative}, WGAN~\cite{arjovsky2017wasserstein}, 
LSGAN~\cite{mao2017least}, WGAN-GP~\cite{gulrajani2017improved}, 
GLANN~\cite{hoshen2019non} and Diffusion-based model \cite{ho2020denoising} 
of resolution $64\times 64$. 
Note that these methods along with the selected resolution are ones
that we could find over the Internet so as to allow a fair comparison to
the best available implementations. Specifically, we take generated
images of GAN, WGAN and LSGAN from celeba-gan-pytorch github\footnote{\url{https://github.com/joeylitalien/celeba-gan-pytorch}}. We take those of WGAN-GP from WGAN-GP-DRAGAN-Celeba-Pytorch github\footnote{\url{https://github.com/joeylitalien/celeba-gan-pytorch}}. For the diffusion model, we take the pre-trained model from pytorch-diffusion-model-celebahq github\footnote{\url{https://github.com/FengNiMa/pytorch_diffusion_model_celebahq}},
which generates samples of resolution $ 256\times 256 $. We resize
generated samples to the resolution of $64\times 64$ to make them comparable
with other methods. Fig.~\ref{fig:CelebaComparisonDL} 
compares generated images by prior DL-based generative models and
our \method{{}} for the CelebA dataset. It can be seen that generated
images of \method{} are comparable with those of prior DL-based methods. 
There are some noise patterns in our results. Their suppression is an
interesting future research topic.

{\bf Objective Evalution.} We use the Frechet Inception Distance (FID)
\cite{heusel2017gans} score to perform quantitative comparison of our
method with prior art.  FID is a commonly used metric to evaluate the
performance of generative models. It considers both diversity and
fidelity of generated images. We follow the procedure described in
\cite{lucic2018gans} to obtain the FID scores; an Inception neural
network extracts features from a set of 10K generated images as well as
another set of 10K real (test) images. Two multivariate Gaussians are
fit to the extracted features from two sets separately. Then, the
Frechet distance between their mean vectors and covariance matrices is
calculated. A smaller FID score is more desirable as it indicates a
better match between the synthesized and real test samples. 

\begin{table}[t]
\caption{Comparison of FID scores for MNIST, Fashion-MNIST and CelebA datasets.}\label{tab:fid_score}
\centering
\begin{tabular}{l|lll}
Method & MNIST  & Fashion  & CelebA  \\ \hline
MM GAN \cite{goodfellow2014generative} & 9.8  & 29.6  & 65.6  \\
NS GAN \cite{goodfellow2014generative} & 6.8  & 26.5  & 55.0  \\
LSGAN \cite{mao2017least} & 7.8  & 30.7  & 53.9  \\
WGAN \cite{arjovsky2017wasserstein} & 6.7 & 21.5   & 41.3  \\
WGAN-GP \cite{gulrajani2017improved} & 20.3  & 24.5  & 30.0  \\
DRAGAN \cite{kodali2017convergence} & 7.6  & 27.7  & 42.3  \\
BEGAN \cite{berthelot2017began} & 13.1  & 22.9  & 38.9  \\ \hline
VAE \cite{kingma2013auto} & 23.8  & 58.7  & 85.7  \\
GLO \cite{bojanowski2017optimizing} & 49.6  & 57.7  & 52.4  \\
GLANN \cite{hoshen2019non} & 8.6  & 13.0  & 46.3 \\
Diffusion \cite{ho2020denoising} & N/A  &N/A  & 48.8 \\ \hline
GenHop \cite{lei2022genhop} & 5.1  & 18.1  & 40.3  \\ \hline
\method{} (Ours) & 9.5  & 19.3  & 43.8 \\ 
\end{tabular}
\end{table}

The FID scores of various methods for MNIST, Fashion-MNIST and CelebA
datasets are compared in Table~\ref{tab:fid_score}.  Methods in the
first and second sections are both based on DL.  Methods in the first
section are adversarial generative models while those in the second
section are non-adversarial. The results of the first and second
sections are taken from \cite{lucic2018gans} and \cite{hoshen2019non},
respectively. For the Diffusion model, we generated 10K
samples using the pre-trained model from pytorch-diffusion-model-celebahq github\footnote{\url{https://github.com/FengNiMa/pytorch_diffusion_model_celebahq}}
and measured the FID score. GenHop in Section 3 does not use a neural
network backbone. Its results are taken from \cite{lei2022genhop}.  We
see from Table~\ref{tab:fid_score} that the FID scores of \method{} are
comparable with those of prior generative models.  In training \method{} model
for Table~\ref{tab:fid_score}, we used 100K training images from CelebA
and 60K training images from MNIST and Fashion-MNIST with no augmentation. 

\method{} is still in its preliminary development stage.
Although it does not outperform prior generative models in the FID
score, it does have comparable performance in all three datasets,
indicating its potential to be further improved in the future. In
addition, \method{} has several other advantages to be discussed in the
next subsection.

\begin{figure*}[t]
\centering
\includegraphics[width=\linewidth]{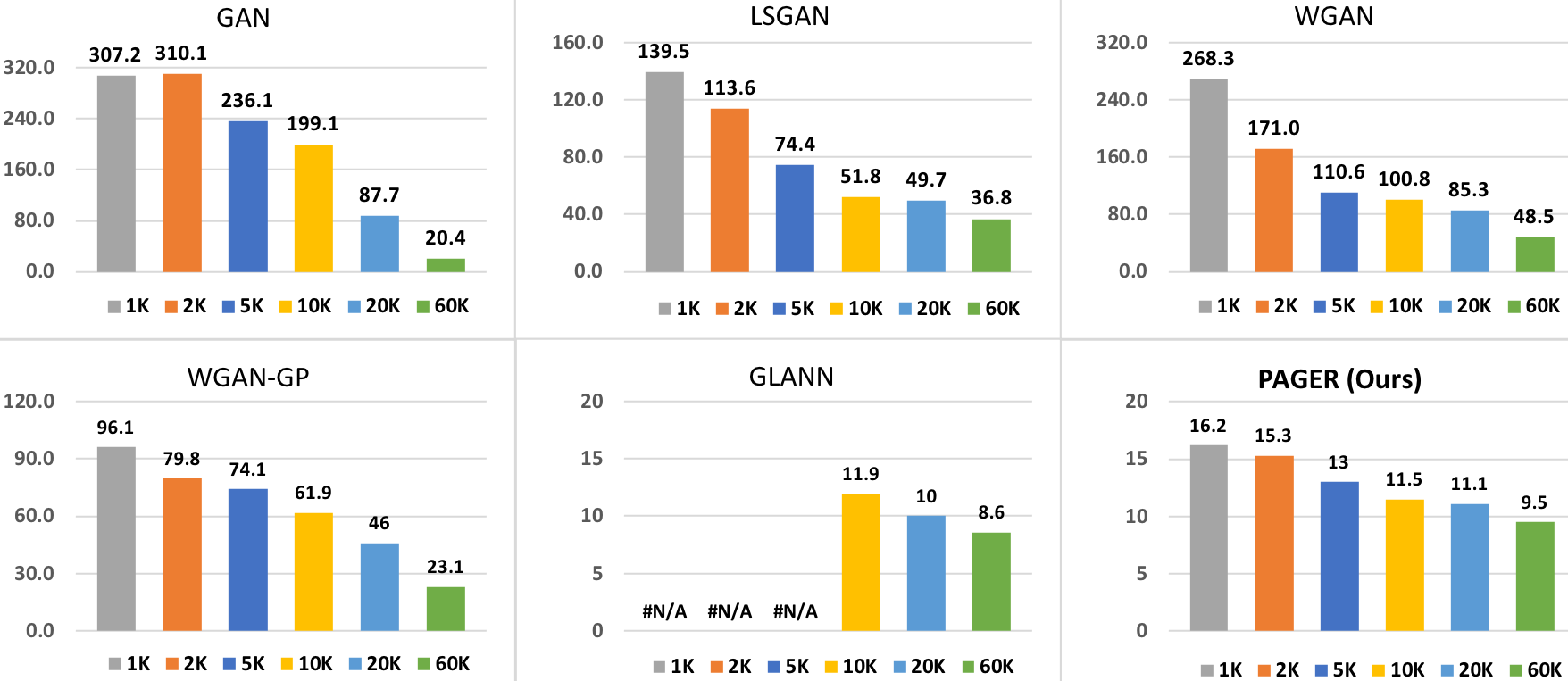}
\caption{Comparison of FID scores of six benchmarking methods with six
training sizes (1K, 2K, 5K, 10K, 20K and 60K) for the MNIST dataset.
The FID scores of \method{} are significantly less sensitive with
respect to smaller training sizes.} \label{fig:training_comparison}
\end{figure*}

\subsection{Other Performance Metrics}

In this section, we study additional performance metrics: robustness to
the number of training samples and training time. 

{\bf Robustness to training dataset sizes.}
Fig.~\ref{fig:training_comparison} presents the FID score of \method{}
and five DL-based generative models (MM GAN, LSGAN, WGAN, WGAN-GP and
GLANN) when the number of training samples is set to 1K, 2K, 5K, 10K, 
20K and 60K for MNIST dataset. To produce the FID scores of the GAN-based related
work, we use the open-source implementation by 
PyTorch-GAN github\footnote{\url{https://github.com/eriklindernoren/PyTorch-GAN}}.
For GLANN, we use the implementation provided by the authors. Since
GLANN is not trained with less than 10K samples, its FID scores for 1K,
2K and 5K samples are not available. It is worth noting that the FID
scores for 60K training samples of some prior work in
Fig.~\ref{fig:training_comparison} are different than those in
Table~\ref{tab:fid_score}. This happens because some of prior generative
models (e.g., MM GAN, LSGAN, and WGAN) are too sensitive to training
hyper-parameters and/or data augmentation \cite{lucic2018gans}. 
The scores reported in Fig.~\ref{fig:training_comparison} are the best
FID scores obtained using the default hyper-parameters in the
open-source library.  We see from Fig.~\ref{fig:training_comparison}
that \method{} is least affected by the number of training samples. Even
with the number of training samples as small as 1K, \method{} has an FID
score of 16.2 which is still better than some prior works' original FID
scores presented in Table \ref{tab:fid_score}, such as WGAN-GP, VAE and
GLO.  Among prior works, GLANN is less sensitive to training size but
cannot be trained with less than 10K samples. 

\begin{table}[t]
\caption{Training time comparison.}\label{tab:runtime}
\centering
\begin{tabular}{l|ll}
Method & CPU  & GPU  \\ \hline
MM GAN \cite{goodfellow2014generative} & 93m14s & 33m17s  \\
LSGAN \cite{mao2017least} & 1426m23s  & 45m52s \\
WGAN \cite{arjovsky2017wasserstein} & 48m11s & 25m55s  \\
WGAN-GP \cite{gulrajani2017improved} & 97m9s  & 34m7s \\ \hline
GLO \cite{bojanowski2017optimizing} & 1090m7s  & 139m18s  \\
GLANN \cite{hoshen2019non} & 1096m24s  & 142m19s \\ \hline
GenHop \cite{lei2022genhop} & 6m12s & N/A \\ \hline
\method{} (Ours) & 4m23s  & 2m59s \\ 
\end{tabular}
\end{table}

\begin{figure}[t]
\centering
\includegraphics[width=0.8\columnwidth]{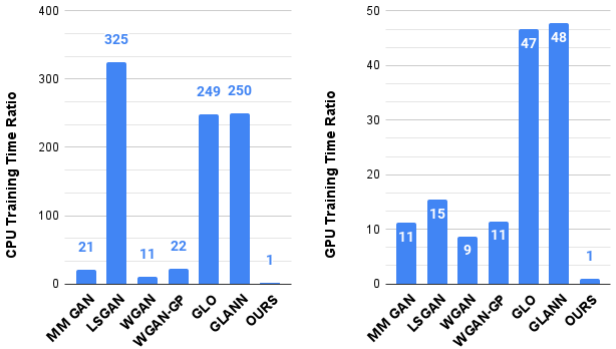}
\caption{Comparison of normalized training time, where each bar represents the training time of a DL-based model corresponding to those shown in Table~\ref{tab:runtime} and normalized by training time of \method{}.} \label{fig:time_ratio}
\end{figure}

{\bf Comparison on Training Time.} The training time of \method{} is
compared with prior work in Table~\ref{tab:runtime} on two platforms. 
\begin{itemize}
\item CPU (Intel Xeon 6130): The CPU training time of \method{} is
slightly more than 4 minutes, which is significantly less than all other methods as shown in Table~\ref{tab:runtime}.  The normalized CPU training times of various DL-based methods against \method{} are visualized in the left subfigure of Fig.~\ref{fig:time_ratio}. \method{} is $11\times$ faster than WGAN and $325\times$ faster than LSGAN. 

\item GPU (NVIDIA Tesla V100): The GPU training time of \method{} is
around 3 minutes, which is again less than all other methods as shown in
Table~\ref{tab:runtime}.  The normalized GPU run times of various
methods are also visualized in the right subfigure of
Fig.~\ref{fig:time_ratio}. \method{} is $9\times$ faster than WGAN and
$48\times$ faster than GLANN. 
\end{itemize}

\begin{figure}[t]
\centering
\includegraphics[width=0.75\columnwidth]{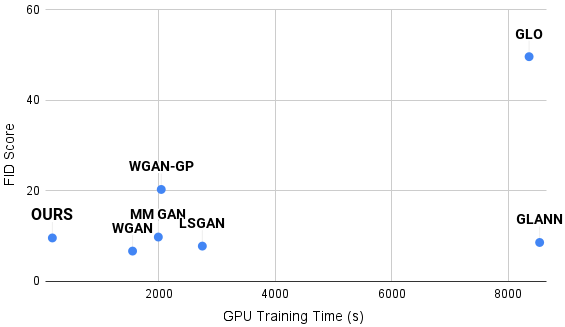}
\caption{Comparison of joint FID scores and GPU training time of \method{} with DL-based related
work in the generation of MNIST-like images. \method{} provides the best
overall performance since it is closest to the left-bottom corner.}
\label{fig:FID_time}
\end{figure}

{\bf Joint Consideration of FID Scores and Training Time.} To provide a
better picture of the tradeoff between training time and FID score, we
present both of these metrics in Fig.~\ref{fig:FID_time}. On this
figure, points that are closer to the bottom left are more desirable. As
seen, \method{} significantly outperforms prior art when considering FID
scores and training time jointly. 

\subsection{Discussion}

\begin{figure}[t]
\centering
\includegraphics[width=\columnwidth]{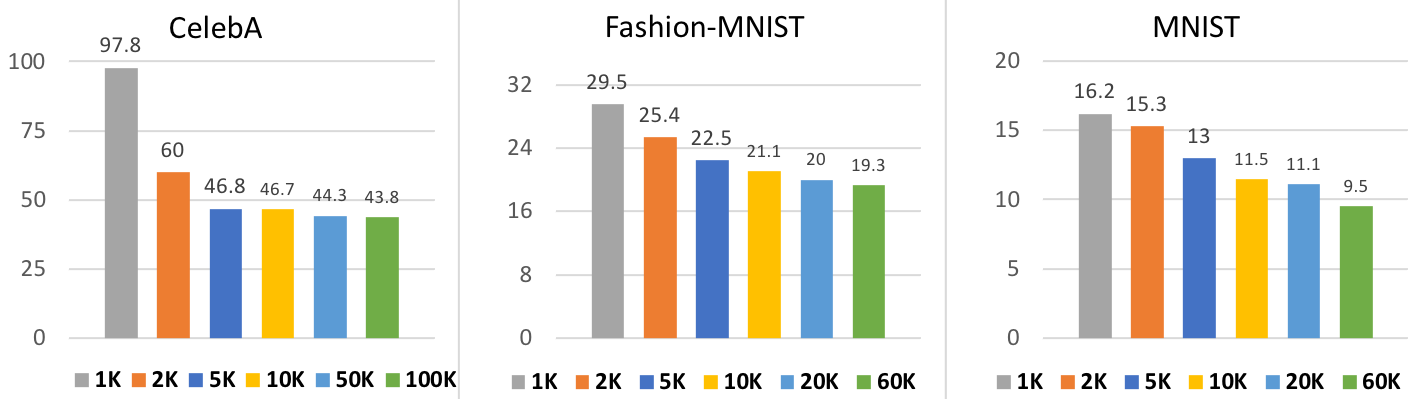}
\caption{Comparison of \method{}'s FID scores with six training
sample sizes for CelebA, Fashion-MNIST and MNIST datasets. We see that
the FID scores do not increase significantly as the training samples
number is as low as 5K for CelebA and 1K for MNIST and Fashion-MNIST.} 
\label{fig:training}
\end{figure}

Based on the above experimental results, we can draw the following
conclusions. 
\begin{itemize}
\item \textbf{Quality image generation.} The FID scores of \method{} are
comparable with those of prior DL-based image generation techniques on
common datasets. This indicates that \method{} can generate images of
similar quality to prior art. 
\item \textbf{Efficient training.} \method{} can be trained in a
fraction of the time required by DL-based techniques. For example, our
MNIST generative model is trained in 4 minutes on a personal computer's
CPU while the fastest prior work demands 25-minute training time on an
industrial GPU.  The efficiency of \method{} is achieved by the
development of a non-iterative training scheme. CPU-based efficient
training implies smaller energy consumption and carbon footprint than
GPU-based DL methods. This is a major advantage of \method{}. 
\item \textbf{Robustness to training sample size.} \method{} can still
yield images of reasonable quality even when the number of training
samples is drastically reduced. For example, in Fig.~\ref{fig:training}
we show that the number of training samples can be reduced from 100K to
5K with only a negligible drop in the generated image quality for the
CelebA dataset. 
\item \textbf{Improvements over prior SSL-based generative
model - GenHop.} While \method{} is the second SSL-based generative
model, it is worthwhile to review its improvements over the prior
SSL-based generative model known as GenHop \cite{lei2022genhop}. First,
the great majority of CelebA generated samples by GenHop suffer from
over-smoothing which blurs details and even fades out the facial
components in many samples as shown in Fig.
\ref{fig:CelebaComparisonGenhop}.  This is because GenHop heavily relies
on LLE which has a smoothing effect and limits synthesis diversity.
On the other hand, \method{} generates diverse samples with visible
facial components. Note that \method{} only uses LLE to add
residuals to already generated samples. It serves as a sharpening
technique and does not affect synthesis diversity. Second, GenHop
limits the resolution of generated samples to $32\times 32$. This
prevents GenHop to be extendable to high-resolution image generation or
other generative applications like super-resolution.  Third, GenHop
takes longer time that \method{} to train and it is not implemented
for GPU training.  Fourth, GenHop only conducts unconditional image
generation while \method{} has further applications such as
attribute-guided image generation and super-resolution.
\end{itemize}

\section{Comments on Extendability}\label{sec:applications}

In this section, we comment on another advantage of \method{}.  That is,
\method{} can be easily tailored to other contexts without re-training.
We elaborate on three applications at the conceptual level. 

\begin{figure*}[t]
\begin{subfigure}{\textwidth}
\includegraphics[width=\linewidth]{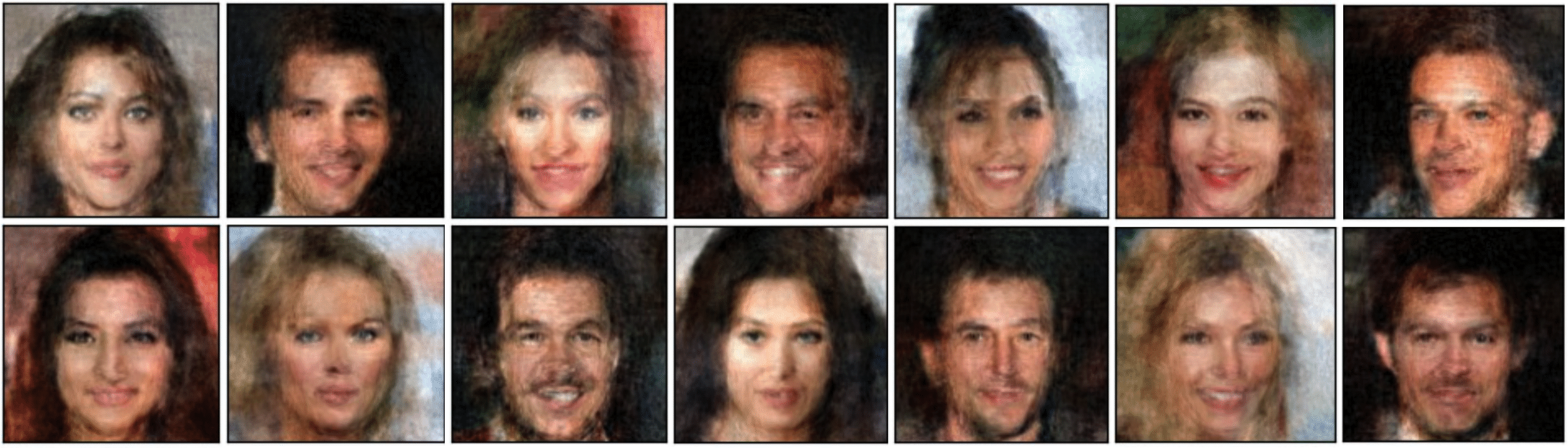}
\caption{Resolution $128\times 128$.}\label{fig:highres128}
\end{subfigure}
\begin{subfigure}{\textwidth}
\includegraphics[width=\linewidth]{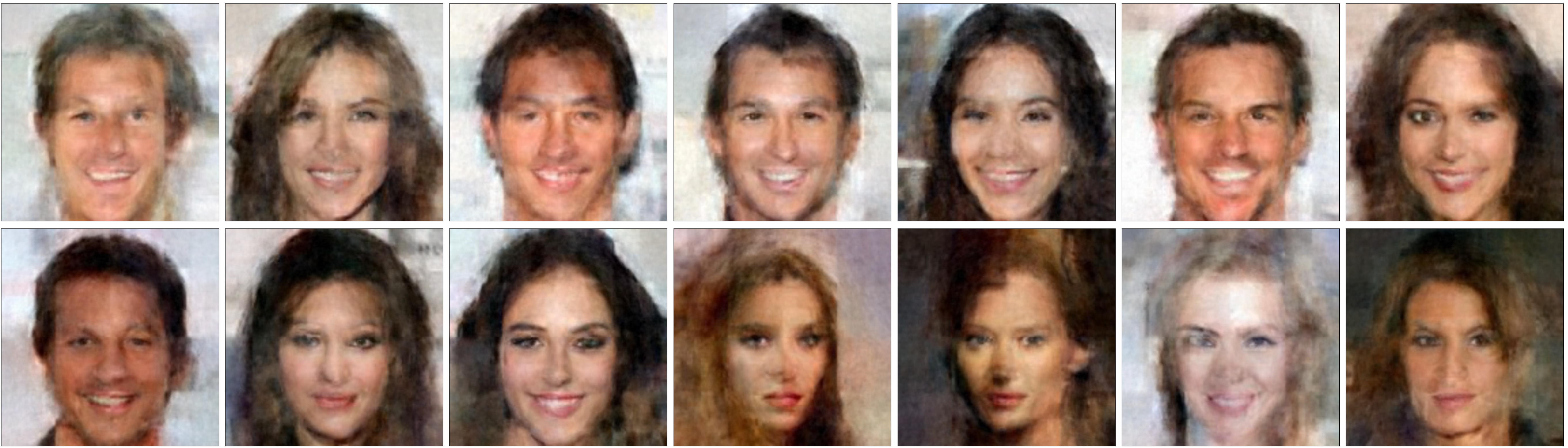}
\caption{Resolution $256\times 256$.}\label{fig:highres256}
\end{subfigure}
 \caption{Examples of generated CelebA-like images of resolution $128\times 128$ and $256\times 256$.}\label{fig:highres}
\end{figure*}

\begin{itemize}

\item \textbf{Super Resolution.} \method{}'s two conditional image
generation modules (i.e., the resolution enhancer and the quality
booster) can be directly used for image super resolution with no
additional training.  These modules enhance the image resolution from an
arbitrary dimension $2^d\times 2^d$ to $2^{d+k}\times 2^{d+k}$, where
$k$ is the number of consecutive resolution enhancer and quality booster
modules needed to achieve this task. Fig.~\ref{fig:superresolution}
shows several examples starting from different resolutions and ending at
resolutions $32\times 32$, $64\times 64$ and $128\times 128$. 

\begin{figure*}[t]
\centering
\includegraphics[width=\linewidth]{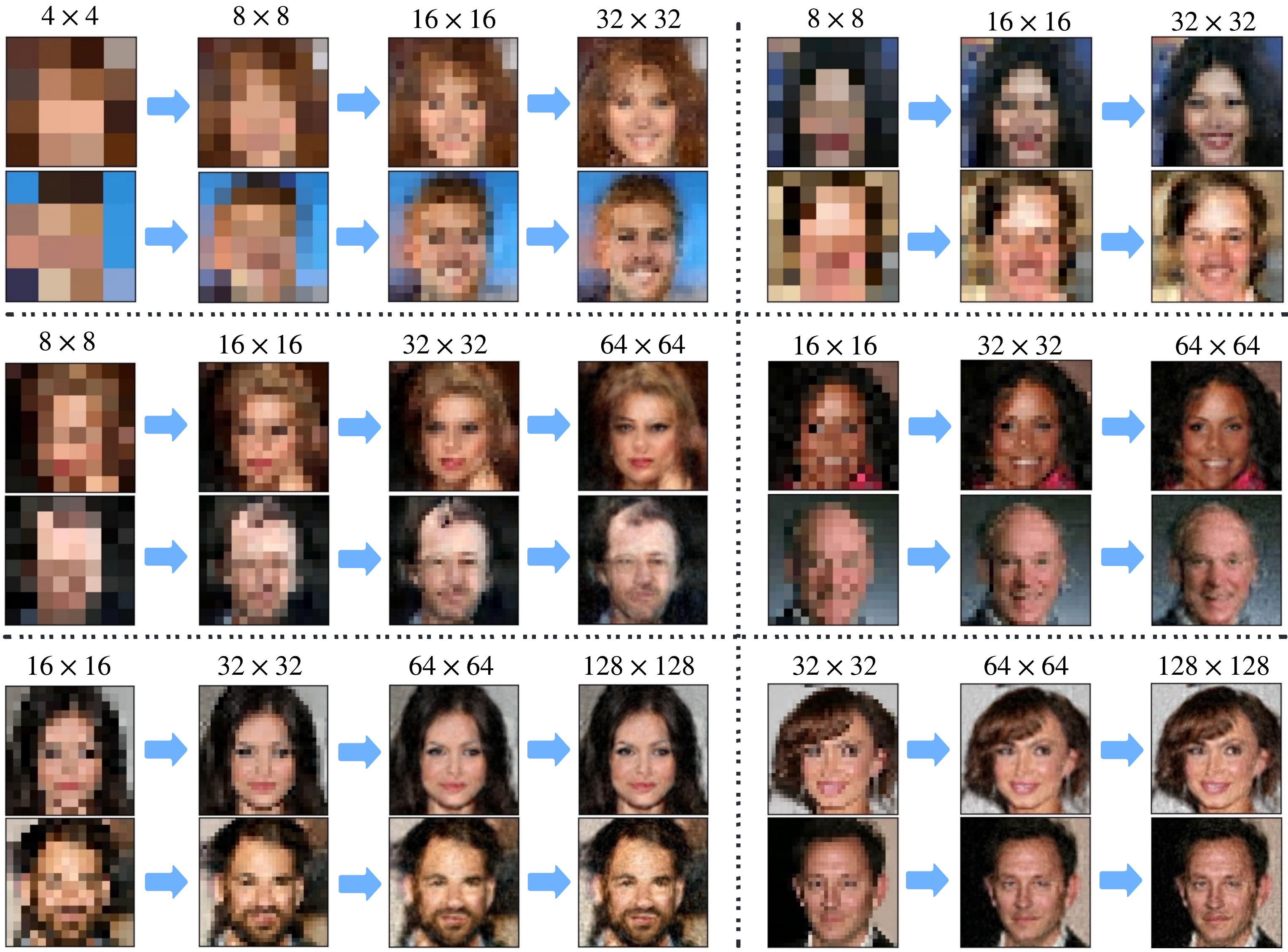}
\caption{Illustration of \method{}'s application in image
super-resolution for CelebA images: Two top rows starting from
resolution $4\times 4$ (left block) and $8\times 8$ (right block) and
ending at resolution $32\times 32$. Two middle rows starting from
resolution $8\times 8$ (left block) and $16\times 16$ (right block) and
ending at resolution $64\times 64$. Two bottom rows starting from
resolution $16\times 16$ (left block) and $32\times 32$ (right block)
and ending at resolution $128\times 128$.}\label{fig:superresolution}
\end{figure*}

\item \textbf{Attribute-guided Face Image Generation.} To generate human
face images with certain characteristics (e.g., a certain gender, hair
color, etc.) we partition the training data based on the underlying
attributes and construct subsets of data (Sec.~\ref{subsec:attributes}).
Each subset is used to train a different core generator that represents
the underlying attributes.  Examples of such attribute-guided face
generation are presented in Figure~\ref{fig:attributes}.  The
feasibility of training \method{} using a subset of training data is a
direct result of its robustness to the training dataset size.  It was
empirically evaluated in Fig.~\ref{fig:training}. The mean FID score of
CelebA-like image generation changes only 6\% when the number of
training samples is reduced from 100K to as low as 5K. 

\item \textbf{High-Resolution Image Generation.} \method{} can be easily
extended to generate images of higher resolution. To achieve this
objective, we can have more resolution enhancer and quality booster
units in cascade to reach the desired resolution. We present several
generated CelebA-like samples of resolution $128\times 128$ and
$256\times 256$ in Fig. \ref{fig:highres}. This gives some evidence that the current design of \method{} is extendable to higher resolution generation. On the other hand, to generate results comparable with state-of-the-art generative models like ProGAN \cite{karras2017progressive}, StyleGAN \cite{karras2019style, karras2020analyzing, karras2021alias}, VQ-VAE-2 \cite{razavi2019generating} or diffusion-based models \cite{dhariwal2021diffusion, ho2022cascaded}, we need to further
optimize our method. Further improvement on \method{} could lead to
enhanced quality of generated images in higher resolutions.

\end{itemize}

\section{Conclusion and Future Work}\label{sec:conclusion}

A non-DL-based generative model for visual data generation called
\method{} was proposed in this work. \method{} adopts the successive
subspace learning framework to extract multi-scale features and learns
unconditional and conditional probability density functions of extracted
features for image generation. The unconditional probability model is
used in the core generator module to generate low-resolution images to
control the model complexity. Two conditional image generation modules,
the resolution enhancer and the quality booster, are used to enhance the
resolution and quality of generated images progressively.  \method{} is
mathematically transparent due to its modular design. We showed that
\method{} can be trained in a fraction of the time required by DL-based
models. We also demonstrated \method{}'s generation quality as the
number of training samples decreases. We then showed the extendibility
of \method{} to image super resolution, attribute-guided face image
generation, and high resolution image generation. 

The model size of \method{} is primarily determined by the sizes
of the quality booster. The number of parameters is about 46 millions.
The large quality booster size is due to the use of LLE in predicting
residual details. We do not optimize the LLE component in the current
implementation. As a future topic, we would like to replace it with a
lightweight counterpart for model size reduction.  For example, We might
replace LLE with GMMs to learn the distribution of residual textures, to
reduce the model size significantly. With these techniques, we aim to
reduce to the model size to less than 10 million parameters.

\section{Acknowledgments}

The authors acknowledge the Center for Advanced Research Computing (CARC) at the University of Southern California for providing computing resources that have contributed to the research results reported within this publication. URL: \url{https://carc.usc.edu}.

\bibliographystyle{unsrt}  
\bibliography{references}  

\end{document}